\documentclass[runningheads]{llncs}
\usepackage{graphicx}
\usepackage{amsmath,amssymb} % define this before the line numbering.
\usepackage{color}
\usepackage[width=122mm,left=12mm,paperwidth=146mm,height=193mm,top=12mm,paperheight=217mm]{geometry}
\begin{document}
% \linenumbers
\pagestyle{headings}
\mainmatter

\title{Learning Temporal Transformations From Time-Lapse Videos} % Replace with your title

\titlerunning{Learning Temporal Transformations From Time-Lapse Videos}

\authorrunning{Yipin Zhou  \space \space Tamara L. Berg}

\author{Yipin Zhou  \space \space Tamara L. Berg}

%Please write out author names in full in the paper, i.e. full given and family names. 
%If any authors have names that can be parsed into FirstName LastName in multiple ways, please include the correct parsing, in a comment to the volume editors:
%\index{Lastnames, Firstnames}
%(Do not uncomment it, because you may introduce extra index items if you do that...)

%\institute{University of North Carolina at Chapel Hill}
\institute{University of North Carolina at Chapel Hill\\
	\email{ \{yipin,tlberg\}@cs.unc.edu}
}

\maketitle

\begin{abstract}
Based on life-long observations of physical, chemical, and biologic phenomena in the natural world, humans can often easily picture in their minds what an object will look like in the future. But, what about computers? In this paper, we learn computational models of object transformations from time-lapse videos. In particular, we explore the use of generative models to create depictions of objects at future times. These models explore several different prediction tasks: generating a future state given a single depiction of an object, generating a future state given two depictions of an object at different times, and generating future states recursively in a recurrent framework. We provide both qualitative and quantitative evaluations of the generated results, and also conduct a human evaluation to compare variations of our models.

\end{abstract}

\section{Introduction}

Before they can speak or understand language, babies have a grasp of some
natural phenomena. They realize that if they drop the spoon it will fall to the
ground (and their parent will pick it up). As they grow older, they develop
understanding of more complex notions like object constancy and time. Children
acquire much of this knowledge by observing and interacting with the world. 

In this paper we seek to learn computational models of how the world works
through observation.  Specifically, we learn models for natural transformations
from video. To enable this learning, we collect time-lapse videos demonstrating
four different natural state transformations: melting, blooming, baking, and
rotting. Several of these transformations are applicable to a variety of
objects. For example, butter, ice cream, and snow melt, bread and pizzas bake,
and many different objects rot. We train models for each transformation --
irrespective of the object undergoing the transformation -- under the
assumption that these transformations have shared underlying physical
properties that can be learned.

To model transformations, we train deep networks to generate depictions of the
future state of objects.  We explore several different generation tasks for
modeling natural transformations.  The first task is to generate the future
state depiction of an object from a single image of the object
(Sec~\ref{pair}).  Here the input is a frame depicting an object at time t, and
output is a generated depiction of the object at time t+k, where k is specified
as a conditional label input to the network.  For this task we explore two
auto-encoder based architectures.  The first architecture is a baseline
algorithm built from a standard auto-encoder framework.  The second architecture is a
generative adversarial network where in addition to the baseline auto-encoder,
a discriminator network is added to encourage more realistic outputs. 

For our second and third future prediction tasks, we introduce different ways
of encoding time in the generation process. In our two-stack model
(Sec~\ref{twostacks}) the input is two images of an object at time t and t+m
and the model learns to generate a future image according to the implicit time
gap between the input images (ie generate a prediction of the object at time
t+2m). These models are trained on images with varying time gaps. 
Finally, in our last prediction task, our goal is to recursively generate
the future states of an object given a single input image of the object. For
this task, we use a recurrent neural network to recursively generate future
depictions (Sec~\ref{recurrent}).  For each of the described future generation
tasks, we also explore the effectiveness of pre-training on a large set of
images, followed by fine-tuning on time-lapse data for improving performance.

Future prediction has been explored in previous works for generating the next
frame or next few frames of a video \cite{LSTM,Yann,Ranzato2014}. Our focus, in comparison,
is to model general natural object transformations and to model future
prediction at a longer time scale (minutes to days) than previous approaches.

We evaluate the generated results of each model both quantitatively and
qualitatively under a variety of different training scenarios (Sec~\ref{exp}).  In 
addition, we perform human evaluations of model variations and image retrieval experiments.  Finally, to help
understand what these models have learned, we also visualize common
motion patterns of the learned transformations. These results are discussed in Sec~\ref{addexp}.

The innovations introduced by our paper include:
1) A new problem of modeling natural object transformations with deep networks,
2) A new dataset of 1463 time-lapse videos depicting 4 common object transformations,
3) Exploration of deep network architectures for modeling and generating future depictions,
4) Quantitative, qualitative, and human evaluations of the generated results, and
5) Visualizations of the learned transformation patterns.

\subsection{Related work}
\label{sec:relatedwork}

\noindent {\bf Object state recognition:} Previous works~\cite{FarhadiCVPR09,attributes,SunAttribute} have looked at the problem of recognizing attributes, which has significant conceptual overlap with the idea of object state recognition. For example ``in full bloom'' could be viewed as an attribute of flowers. Parikh and Grauman~\cite{attributes} train models to recognize the relative strength of attributes such as face A is ``smiling more'' than face B from ordered sets of images. One way to view our work is as providing methods to train relative state models in the temporal domain. 
Most relevant to our work, given a set of object transformation terms, such as ripe vs unripe, \cite{stateCVPR2015} learns visual classification and regression models for object transformations from a collection of photos. In contrast, our work takes a deep learning approach and learns transformations from video -- perhaps a more natural input format for learning temporal information.\\
\noindent {\bf Timelapse data analysis:} Timelapse data captures changes in
time and has been used for various applications. \cite{Shih} hallucinates an
input scene image at a different time of day by making use of a timelapse
video dataset exhibiting lighting changes in an example-based color transfer
technique. \cite{steven1} presents an algorithm that synthesizes timelapse
videos of landmarks from large internet image collections. In their follow-up
work,~\cite{steven2} imports additional camera motion while composing videos
to create transformations in time and space.\\
\noindent {\bf Future prediction:} Future prediction has been applied to various tasks such as estimating the future trajectories of cars\cite{Walker},
pedestrians\cite{Kitani}, or general objects\cite{Yuen} in images or videos. In
the ego-centric activity domain, \cite{pred} encodes the prediction problem as
a binary task of selecting which of two video clips is first in temporal
ordering. Given large amounts of unlabeled video data from the internet,
\cite{carl} trains a deep network to predict visual representations of future
images, enabling them to anticipate both actions and objects before they
appear.\\
\noindent {\bf Image/Frame generation:}  Generative models have attracted
extensive attention in machine
learning~\cite{Hinton1986,Smolensky1986,Lee2009,Hinton2006,Tang2013,Kingma2013}.
Recently many works have focused on generating novel natural or high-quality
images. \cite{chairs} applies deep structure networks trained on synthetic data
to generate 3D chairs. \cite{DRAW} combines
variational auto-encoders with an attention mechanism to recurrently generate
different parts of a single image. Generative adversarial networks (GANs) have
shown great promise for improving image generation quality~\cite{GANs}. GANs
are composed of two parts, a generative model and a discriminative model, to be trained jointly. 
Some extensions have combined GAN structure with
multi-scale laplacian pyramid to produce high-resolution generation
results~\cite{laplacian}. Recently  \cite{DCGAN} incorporated deep
convolutional neural network structures into GANs to improve image quality. \cite{CVPR16context} proposed a network to generate the contents of an arbitrary image region according to its surroundings. Some related approaches~\cite{LSTM,Yann,Ranzato2014} have trained generation models to
reconstruct input video frames and/or generate the next few consecutive
frames of a video. We also explore the use of DCGANs for future prediction, focusing
on modeling object transformations over relatively long time scales.

% The motivation is that they claim the better the ability of model doing reconstruction/prediction, the better internal representation the model has incorporate about the video which can lead to the better feature representation of the video finally.

% table for dataset statistics
\begin{table} [t]
{\small
\begin{center}
\scalebox{1.0}{
\begin{tabular}{|l | l | l | l |}
\hline
Rotting: 185& Melting: 453 & Baking: 242& Blooming: 583\\
\hline\hline
Strawberry: 35 & Ice cream: 128 & Cookies: 55 & Flower: 583\\
Watermelon: 9 & Chocolate: 18  & Bread: 57   & \\
Tomato: 25     & Butter: 9    & Pizza: 59              &  \\
Banana: 26      & Snow: 54    & Cake: 48    &  \\
Apple: 23        & Wax: 60      & Other: 23    &   \\
Peach: 8          & Ice: 184    &      &    \\
Other: 59        &     &    &            \\
\hline
\end{tabular}
} 
\end{center}
}
\caption{Statistics of our transformation categories. Some categories contain multiple objects (e.g. ice cream, chocolate, etc melting) while others apply only to a specific object (e.g. flowers blooming). Values indicate the total number of videos collected for each category.}
\label{table:dataset_stat}
%\vspace{-.5cm}
\end{table}

%\vspace{-.1cm}
\section{Object-centric timelapse dataset}
%\vspace{-.1cm}

Given the high-level goal of understanding temporal transformations of objects, we require a collection of videos showing temporal state changes. Therefore, we collect a large set of object-centric timelapse videos from the web. Time-lapse videos are ideal for our purposes since they are designed to show an entire transformation (or a portion of a transformation) within a short period of time. 

% Since we want to explore different stages of objects and predict the future states based on their transformation, object-centric timelapes videos are good source for those information. Because they show the procedure of such transformation of a single or several objects which used to be a slow process within a short period of time. 

%\vspace{-.1cm}
\subsection{Data collection}
%\vspace{-.1cm}
We observe that time-lapse photography is quite popular due to the relative ease of data collection. A search on YouTube for ``time lapse'' results in over 11 million results. Anyone with a personal camera, GoPro, or even a cell phone can capture a simple time-lapse video and many people post and publicly share these videos on the web. We collect our object-based timelapse video dataset by directly querying keywords through the YouTube API. For this paper, we query 4 state transformation categories: Blooming, Melting, Baking and Rotting, combined with various object categories. This results in a dataset of more than 5000 videos. This dataset could be extended to a wider variety of transformations or to more complex multi-object transformations, but as a first step we focus on these 4 as our initial goal set of transformations for learning.

\begin{figure}[t]
\centering
\includegraphics[width=0.95\textwidth]{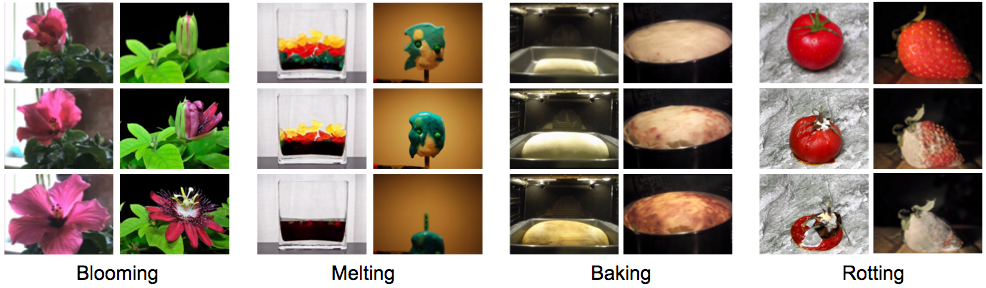}
\caption{Example frames from our dataset of each transformation category: Blooming, Melting, Baking, and Rotting. In each column, time increases as you move down the column, showing how an object transforms.}
%\vspace{-.5cm}
\label{fig:frames}
\end{figure}

For ease of learning, ideally these videos should be object-centric with a static camera capturing an entire object transformation. However, many videos in the initial data collection stage do not meet these requirements. Therefore, we clean the data using Amazon Mechanical Turk (AMT) as a crowdsourcing platform. Each video is examined by 3 Turkers who are asked questions related to video quality. Videos that are not timelapse, contain severe camera motion or are not consistent with the query labels are removed. We also manually adjust parts of the videos which are playing backwards (a technique used in some time-lapse videos), contain more than one round of the specified transformation, and remove irrelevant prolog/epilog. Finally our resulting dataset contains 1463 high quality object based timelapse videos. Table.~\ref{table:dataset_stat} shows the statistics of transformation categories and their respective object counts. Fig.~\ref{fig:frames} shows example frames of each transformation category.

% frame examples figure

%\vspace{-.1cm}
\subsection{Transformation degree annotation}
%\vspace{-.1cm}
To learn natural transformation models of objects from videos, we need to first
label the degree of transformation throughout the videos. In language, people
use text labels to describe different states, for instance, fresh vs rotted
apple. However, the transformation from one state to another is really a
continuous evolution.  Therefore, we represent the degree of transformation
with a real number, assigning the start state a value of 0 (not at all rotten)
and the end state (completely rotten) a value of 1. To annotate objects from
different videos we could naively assign the first frame a value of 0 and the
last frame a value of 1, interpolating in between.  However, we observe that
some time-lapse videos may not depict entire transformations, resulting in poor
alignments.

Therefore, we design a labeling task for people to assign degrees of
transformation to videos. Directly estimating a real value for frames turns out
to be impractical as people may have different conceptions of transformation
degree.  Instead our interface displays reference frames of an object
category-transformation and asks Turkers to align frames from a target video to
the reference frames according to degree of transformation.  Specifically, for
each object category-transformation pair we select 5 reference frames from a
reference video showing: transformation degree values of 0, 0.25, 0.5, 0.75,
and 1. Then, for the rest of the videos in that object-transformation category,
we ask Turkers to select frames visually displaying the same degree of
transformation as the reference frames. If the displayed video does not depict
an entire transformation, Turkers may align less than 5 frames with the
reference. Each target video is aligned by 3 Turkers and the median of their
responses is used as annotation labels(linearly interpolating degrees between
labeled target frames).  This provides us with consistent degree annotations
between videos.

\begin{figure} [t]
\centering
\includegraphics[width=0.95\textwidth]{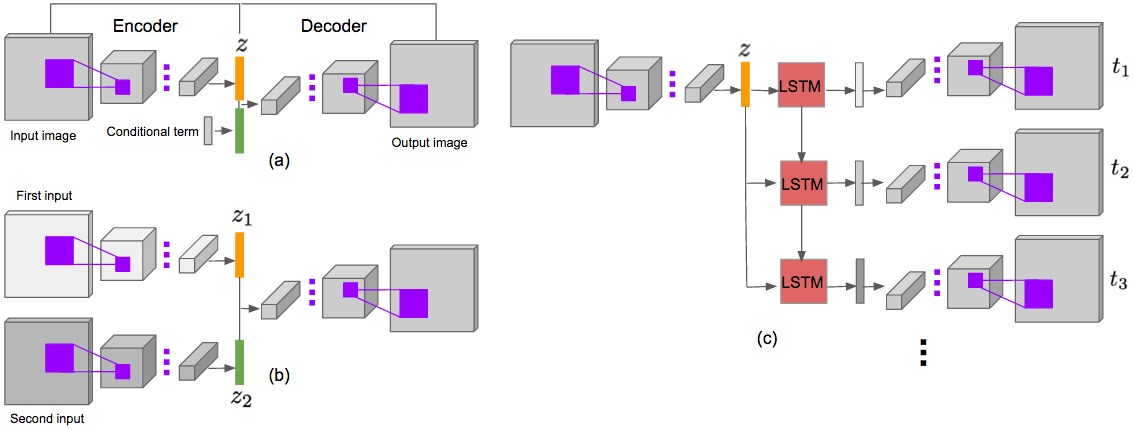}
\caption{Model architectures of three generation tasks: (a) Pairwise generator; (b) Two stack generator; (c) Recurrent generator.}
%\vspace{-.25cm}
\label{fig:model_arch}
\end{figure}

%\vspace{-.2cm}
\section{Future state generation Tasks \& Approaches}
\label{sec:tasksapproaches}
%\vspace{-.1cm}

Our goal is to generate depictions of the future state of objects. In this work, we explore frameworks for 3 temporal prediction tasks. In the first task (Sec~\ref{pair}), called pairwise generation, we input an object-centric frame and the model generates an image showing the future state of this object. Here the degree of the future state transformation -- how far in the future we want the depiction to show -- is controlled by a conditional term in the model. In the second task (Sec~\ref{twostacks}) we have two inputs to the model: two frames from the same video showing a state transformation at two points in time. The goal of this model is to generate a third image that continues the trend of the transformation depicted by the first and second input. We call this ``two stack'' generation. In the third task (Sec~\ref{recurrent}), called ``recurrent generation'', the input is a single frame and the goal is to recursively generate images that exhibit future degrees of the transformation in a recurrent model framework.

%\vspace{-.1cm}
\subsection{Pairwise generation}
\label{pair}
%\vspace{-.1cm}
In this task, we input a frame and generate an image showing the future
state. We model the task using an autoencoder style convolutional neural
network. The model architecture is shown in Fig.~\ref{fig:model_arch}(a), where
the input and output size are 64x64 with 3 channels and encoding and decoding
parts are symmetric. Both encoding and decoding parts consists of 4
convolution/deconvolution layers with the kernel size 5x5 and a stride of 2,
meaning that at each layer the height and width of the output feature map
decreases/increases by a factor of 2 with 64, 128, 256, 512/512, 256, 128, 64
channels respectively. Each conv/deconv layer, except the last layer, is
followed by a batch normalization operation\cite{BatchNorm} and ReLU activation
function\cite{relu}. For the last layer we use Tanh as the activation function.
The size of the hidden variable $z$ (center) is 512. We represent the
conditional term encoding the degree of elapsed time between the input and
output as a 4 dimensional one-hot vector, representing 4 possible degrees. This
is connected with a linear layer (512) and concatenated
with $z$ to adjust the degree of the future depiction. Below we describe
experiments with different loss functions and training approaches for this
network.

\noindent {\bf p\_mse:} As a baseline, we use pixel-wised mean square error(p\_MSE) between prediction output and ground truth as the loss function. Previous image generation works\cite{LSTM,Yann,SSIM,Ranzato2014,CVPR16context} postulate that pixel-wised $l_2$ criterion is one cause of output generation blurriness. We also observe this effect in the outputs produced by this baseline model.

% Specifically, while training besides the autoencoder or we call generator (G) we also import a CNN binary classifier as a discriminator (D), which also incorporating the conditional term by connecting one-hot vector with a linear layer and reshaping to the same size with the input images and concatenating to it's third dimensions. 
% The architecture of D is the same with the encoder of G, except that the last output is a scalar and connect with sigmoid function. The two parts G and D are trained simultaneously. For D, it attempts to discriminate generated images from G and real images from ground truth.

\noindent {\bf p\_mse+adv:} Following the success of recent image generation works\cite{GANs,DCGAN}, which make use of generative adversarial networks (GANs) to generate better quality images, we explore the use of this network architecture for temporal generation. For their original purpose, these networks generate images from randomly sampled noise. Here, we use a GAN to generate future images from an image of an object at the current time. Specifically, we use the baseline autoencoder previously described and incorporate an adversarial loss in combination with the pixel-wise MSE. During training 
this means that in addition to the auto-encoder, called the generator (G), we also train a binary CNN
classifier, called the discriminator (D). The discriminator takes as input, the output of the generator and is trained to classify images as real or fake (i.e. generated). These two networks are adversaries because G is trying to generate images that can fool D into thinking they are real, and D is trying to determine if the images generated by G are real. Adding D helps to train a better generator G that produces more realistic future state images. The architecture of D is the same as the encoder of G, except that the last output is a scalar and connect with sigmoid function. D also incorporates the conditional term by connecting a one-hot vector with a linear layer and reshaping to the same size as the input image then concatenating in the third dimension.
The output of D is a probability, which will be large if the input is a real image and small if the input is a generated image. For this framework, the loss is formulated as a combination of the MSE and adversarial loss: 
\begin{align}
  &L_G = L_{p\_mse} + \lambda_{adv} * L_{adv} \;&
\end{align}
where $L_{p\_mse}$ is the mean square error loss, where $x$ is the input image, $c$ is the conditional term, G(.) is the output of the generation model, and $y$ is the ground truth future image at time = current time + $c$.
\begin{align}
  & L_{p\_mse}  = | y - G(x, c) |^2 \;&
\end{align}
And, $L_{adv}$ is a binary cross-entropy loss with $\alpha$ = 1 that penalizes if the generated image does not look like a real image. D[.] is the scalar probability output by the discriminator.
\begin{align}
  &L_{adv}  = -\alpha\log(D[G(x,c), c]) - (1-\alpha)\log(1- D[G(x,c), c]) \;&
\end{align}
During training, the binary cross-entropy loss is used to train D on both real and generated images. For details of jointly training adversarial structures, please refer to~\cite{GANs}. 

\noindent {\bf p\_g\_mse+adv:} Inspired by \cite{Yann}, where they introduce gradient based $l_1$ or $l_2$ loss to sharpen generated images. We also evaluate a loss function that is a combination of pixel-wise MSE, gradient based MSE, and adversarial loss: 
\begin{align}
  &L_G = L_{p\_mse} + L_{g\_mse}  + \lambda_{adv} * L_{adv} \;&
\end{align}
where $L_{g\_mse}$ represents mean square error in the gradient domain defined as:
\begin{align}
  &L_{g\_mse}  =  ( g_x[y] - g_x[G(x, c)] )^2 + ( g_y[y] - g_y[G(x, c)] )^2\;&
\end{align}
$g_x[.]$ and $g_y[.]$ are the gradient operations along the x and y axis of images. We could apply different weights to $L_{p\_mse} $ and $L_{g\_mse}$, but in this work we simply weight them equally.

% We observe that state transformation is a gradual procedure. Even though the appearance of objects are changing, this still look like their previous state to some degree.

\noindent {\bf p\_g\_mse+adv+ft:}  Since we have limited training data for each transformation, we investigate the use of pre-training for improving generation quality. In this method we use the same loss function as the last method, but instead of training the adversarial network from scratch, training proceeds in two stages. The first stage is a reconstruction stage where we train the generation model using random static images for the reconstruction task. Here the goal is for the output image to match the input image as well as possible even though passing through a bottleneck during generation. In the second stage, the fine-tuning stage, we fine-tune the network for the temporal generation task using our timelapse data. By first pre-training for the  reconstruction task, we expect the network to obtain a good initialization, capturing a representation that will be useful for kick-starting the temporal generation fine-tuning.

%\vspace{-.1cm}
\subsection{Two stack generation}
\label{twostacks}
%\vspace{-.1cm}
In this scenario, we want to generate an image that shows the future state of
an object given two input images showing the object in two stages of its
transformation. The output should continue the transformation pattern
demonstrated in the input images, i.e. if the input images depict the object at
time t and t+m, then the output should depict the object at time t+2m. We
design the generation model using two stacks in the encoding part of the model
as shown in Fig.~\ref{fig:model_arch}(b). The structures of the two stacks are
the same and are also identical to the encoding part of the pairwise generation
model. The hidden variables $z_1$ and $z_2$ are both 512 dimensions, and are
concatenated together and fed into the decoding part, which is also the same as
the previous pairwise generation model. The two stacks are sequential, trained
independently without shared weights. 

Given the blurry results of the baseline for pairwise generation,
here we only use three methods {\bf p\_mse+adv}, {\bf p\_g\_mse+adv}, and
{\bf p\_g\_mse+adv+ft}. The structure of the discriminator is the same. For the
fine-tuning method, during the reconstruction training, we make the two inputs
the same static image. The optimization procedures are the same as for the
pairwise generation task, but we do not have conditional term here (since time
for the future generation is implicitly specified by the input images). 

%\vspace{-.1cm}
\subsection{Recurrent generation}
\label{recurrent}
%\vspace{-.1cm}
In this scenario, we would like to recursively generate future
images of an object given only a single image of its current temporal state.
In particular, we use a recurrent neural network framework for generation where each time step generates an image of the object at a fixed degree interval in the future. This model structure is shown in Fig.~\ref{fig:model_arch}(c). After hidden variable $z$, we add a LSTM\cite{LSTM1997} layer. For each time step, the LSTM layer takes both $z$ and the output from the previous time step as inputs and sends a 512 dimension vector to the decoder. The structure of the  encoder and decoder are the same as in the previous scenarios, where the decoding portions for each time slot share the same weights. 

We evaluate three loss functions in this network: {\bf p\_mse+adv}, {\bf p\_g\_mse+adv} and {\bf p\_g\_mse+adv+ft}. The structure of the discriminator is again the same without the conditional term (as in the two-stack model). For fine-tuning, during reconstruction training, we train the model to recurrently output the same static image as the input at each time step. 

%\vspace{-.2cm}
\section{Experiments}
\label{exp}
%\vspace{-.1cm}

In this section, we discuss the training process and parameter settings for all
experiments (Sec~\ref{params}). Then, we describe dataset pre-processing and
augmentation (Sec~\ref{dataaug}). Finally, we discuss quantitative and
qualitative analysis of results for: pairwise generation (Sec~\ref{pairexp}),
two-stack generation (Sec~\ref{twostacksexp}), and recurrent generation
(Sec~\ref{recurrentexp}).

%\vspace{-.1cm}
\subsection{Training \& Parameter settings}
\label{params}
%\vspace{-.1cm}

Unless otherwise specified training and parameter setting details are applied
to all models. During training, we apply Adam Stochastic
Optimization\cite{Adam} with learning rate 0.0002 and minibatch of size 64. The
models are implemented using the Tensorflow deep learning
toolbox\cite{tensorflow}. In the loss functions where we combine mean square
error loss with adversarial loss (as in equation(1)), we set the weight of the
adversarial loss to $\lambda_{adv}$ = 0.2 for all experiments.

%\vspace{-.1cm}
\subsection{Dataset preprocessing \& augmentation}
\label{dataaug}
%\vspace{-.1cm}

\noindent {\bf Timelapse dataset:} Some of the collected videos depict more
than one object or the object is not located in the center of the frames. In
order to help the model concentrate on learning the transformation itself
rather object localization, for each video in the dataset, we obtain a
cropped version of the frames centered on the main object. 
%We use these cropped versions in all of our experiments.
We randomly split the videos into training and testing sets in a ratio of 0.85
: 0.15. Then, we sample frame pairs (for pairwise generation) or groups of
frames (for two-stack and recurrent generation) from the training and testing
videos. Frames are resized to 64x64 for generation. To prevent overfitting, we
also perform data augmentation on training pairs or groups by incorporating frame crops and left-right flipping.

\noindent {\bf Reconstruction dataset:} This dataset contains static images
used to pre-train our models on reconstruction tasks. Initially, we tried
training only on objects depicted in the timelapse videos and observed performance improvement.
However, collecting images of
specific object categories is a tedious task at large-scale.
Therefore, we also tried pre-training on random images
(scene images or images of random objects) and found that the results were
competitive. This implies that the content of the images is not as important as
encouraging the networks to learn how to reconstruct arbitrary input images
well. We randomly download 50101 images from ImageNet\cite{imagenet} as our
reconstruction dataset. The advantage of this strategy is that we are able to
use the same group of images for every transformation model and task. 
%No data augmentation is applied to this data.

% We make the number of such frame pairs equal to the number of pairs with 0.25 degree interval.

\begin{figure} [t]
\centering
\includegraphics[width=1.0\textwidth]{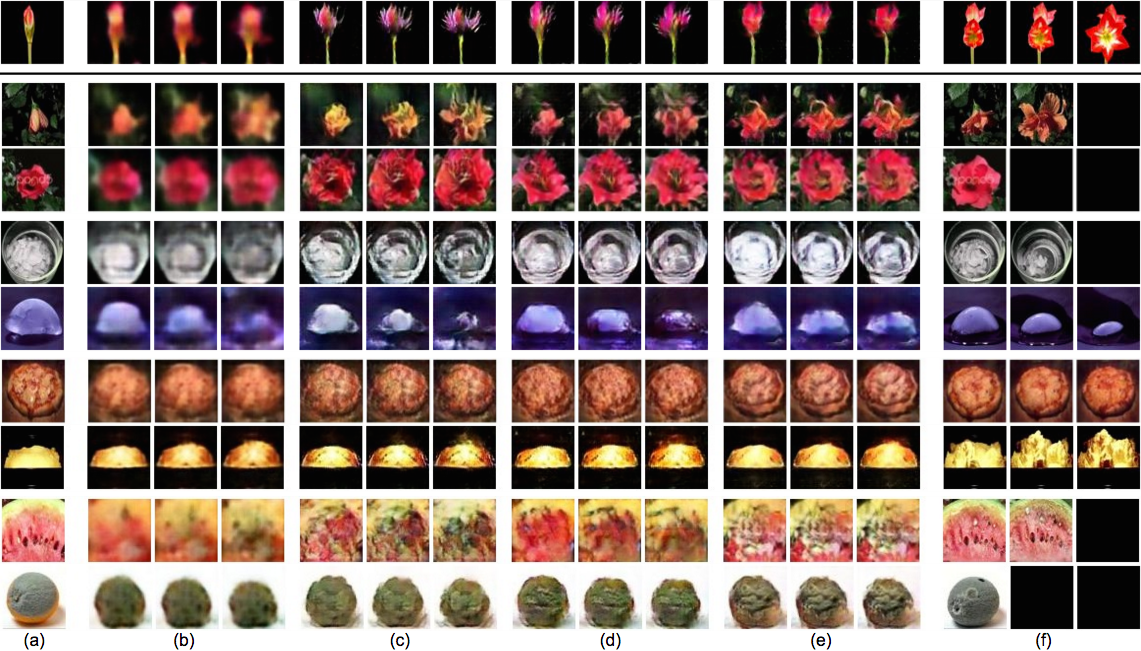}
\caption{Pairwise generation results for: Blooming (rows 1-3), Melting (rows 4-5), Baking (rows 6-7) and Rotting (rows 8-9). Input(a), p\_mse(b),  p\_mse+adv(c), p\_g\_mse+adv(d), p\_g\_mse+adv+ft(e), Ground truth frames(f). Black frames in the ground truth indicate video did not depict transformation that far in the future. }
%\vspace{-.25cm}
\label{fig:pair_gen}
\end{figure}

% table for reconstruction eval
\begin{table} [t]
\small{
\begin{center}
\begin{tabular}{|l |c |c| c || c| c| c ||c| c| c|}
%\hline
\cline{2-10}
\multicolumn{1}{c|}{}&\multicolumn{3}{c||}{Pairwise}&\multicolumn{3}{c||}{Two stack}&\multicolumn{3}{c|}{Recurrent}\\
\cline{2-10}
\multicolumn{1}{c|}{}& PSNR & SSIM & MSE & PSNR & SSIM & MSE & PSNR & SSIM & MSE\\

\hline

p\_mse+adv  & 17.0409 & 0.5576 & 0.0227 & 17.7425 & 0.5970 & 0.0185 & 17.2758 & 0.5747 & 0.0211\\
p\_g\_mse+adv & 17.0660 & 0.5720 & 0.0224 & 17.9157 & 0.6122 & 0.0177 & 17.2951 & 0.5749 & 0.0214\\
p\_g\_mse+adv+ft & 17.4784 & 0.6036 & 0.0207 & 18.6812 & 0.6566 & 0.0153 & 18.3357 & 0.6283 & 0.0166
\\

\hline
\end{tabular}
\end{center}
}
\caption{Quantitative Evaluation of: Pairwise generation, Two stack generation, and Recurrent tasks. For PSNR and SSIM larger is better, while for MSE lower is better.}
\label{table:reconstruction}
%\vspace{-.5cm}
\end{table}

%\vspace{-.1cm}
\subsection{Pairwise generation}
\label{pairexp}
%\vspace{-.1cm}
In the pairwise generation task, we input an image of an object and the model 
outputs depiction of the future state of the object, where the degree of 
transformation is controlled by a conditional term. The conditional term is a 4
dimensional one-hot vector which indicates whether the predicted output should
be 0, 0.25, 0.5 or 0.75 degrees in the future from the input frame. We sample
frame pairs from timelapse videos based on annotated degree value intervals. A
0 degree interval means that the input and output training images are
identical. We consider 0 degree pairs as auxiliary pairs, useful for two
reasons: 1) They help augment the training data with video frames (which display
different properties from still images), 
and 2) The prediction quality of image reconstruction is highly correlated with the quality of
future generation.  Pairs from 0 degree transformations can be easily 
be evaluated in terms of reconstruction quality since the prediction should ideally
exactly match the ground truth image.
Predictions for future degree transformations are somewhat
more subjective. For example, from a bud the resulting generated bloom
may look like a perfectly valid flower, but may not match the exact
flower shape that this particular bud grew into (an example is shown in
Fig.~\ref{fig:pair_gen} row 1).

%which is necessary because even
%with the data augmentation as described above, augmentation with frames from
%videos can be helpful (still images display somewhat different properties from
%video frames); 

% And what this step do is reconstructing the input frame, we can easily quantitatively evaluate the reconstruction results in order to infer the prediction results; We can also directly compare the future prediction results with the `ground truth' future frames from that video. But the problem is that future state can have many possibilities, the outputs can have different orientation or different way of transformation with the `ground truth' but they are still valid prediction. For instance, both turning black while dehydrating and affecting with mildew can be considered as rotting transformation. But our current image comparison algorithms will give a low score for this situation. 

We train pairwise generation models separately for each of the 4 transformation
categories using {\bf p\_mse}, {\bf p\_mse+adv} and {\bf p\_g\_mse+adv}
methods, trained for 12500 iterations on timelapse data from scratch. For the
{\bf p\_g\_mse+adv+ft} method, the models are first trained on the
reconstruction dataset for 5000 iterations with the conditional term fixed as
`0 degree' and then fine-tuned on timelapse data for another 5500 iterations.
We observe that the fine-tuning training converges faster than training from
scratch (example results with 3 different degree conditional terms 
in Fig.~\ref{fig:pair_gen}).  We observe that the
baseline suffers from a high degree of blurriness. Incorporating other terms
into the loss function improves results, as does pre-training with fine-tuning.
Table.~\ref{table:reconstruction} (cols 2-4) shows evaluations of
pairwise-generation reconstruction. For evaluation, we compute the Peak Signal to Noise Ratio(PSNR), Structural Similarity Index(SSIM)
and Mean Square Error (MSE) values between the output and ground truth. We can
see that incorporating gradient loss slightly improves results while
pre-training further improves performance.  This agrees with the qualitative
visual results. 

% \textbf{(Maybe need to explain why PSNR/SSIM scores are not that high and something about baseline...?)}

%Similar to pairwise generation, triples with 0 degree interval are identical
%frames used to augment training data and for quantitative evaluation. 

\begin{figure} [t]
\centering
\includegraphics[width=1.0\textwidth]{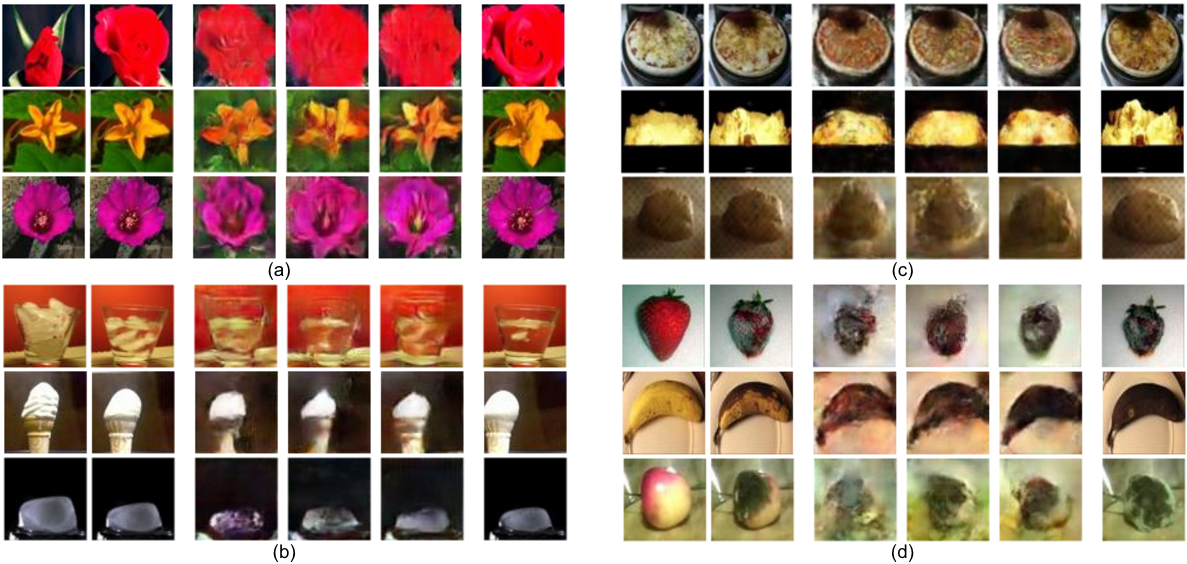}
\caption{Two stack generation results for Blooming (a), Melting (b), Baking (c) and Rotting (d). For each we show the two input frames (col 1-2),  and results for: p\_mse+adv (col 3), p\_g\_mse+adv (col 4), p\_g\_mse+adv+ft (col 5) and ground truth (col 6) }
\label{fig:two_stack_gen_1}
\end{figure}

\begin{figure} [t]
\centering
\includegraphics[width=1.0\textwidth]{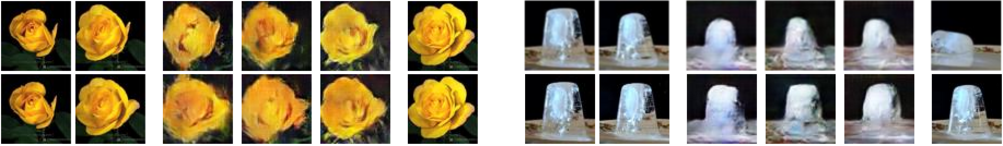}
\caption{Two stack generation with varying time between input images. For each example we show: the two input images (col 1-2),  p\_mse+adv (col 3), p\_g\_mse+adv (col 4), p\_g\_mse+adv+ft (col 5) and ground truth (col 6). The models are able to vary their outputs depending on elapsed time between inputs.}
\label{fig:two_stack_gen_2}
\end{figure}

%\vspace{-.1cm}
\subsection{Two stack generation}
\label{twostacksexp}
%\vspace{-.1cm}
For two stack generation, the model generates an image showing the future
state of an object given two input depictions at different stages of
transformation. As training data, we sample frame triples from videos with
neighboring frame degree intervals: 0, 0.1, 0.2, 0.3, 0.4, and 0.5.
We train two stack generation models for each of the 4 transformation
categories, trained for 12500 iterations for the {\bf p\_mse+adv} and {\bf
p\_g\_mse+adv} methods. For the {\bf p\_g\_mse+adv+ft} method, the models are
first pre-trained on the reconstruction dataset for 4000 iterations and then
fine-tuned on timelapse data for 6500 iterations
(Fig.~\ref{fig:two_stack_gen_1} shows example prediction results).  We observe
that {\bf p\_g\_mse+adv+ft} generates improved results in terms of both image
quality and future state prediction accuracy. We further evaluate the
reconstruction accuracy off these models in Table.~\ref{table:reconstruction}
(cols 5-7). Furthermore, in this task we expect that the models can not only
predict the future state, but also learn to generate the correct time interval
based on the input images.  Fig.~\ref{fig:two_stack_gen_2} shows input images
with different amounts of elapsed time. We can see that the models are able to
vary how far in the future to generate based on the input image interval.

\begin{figure} [t]
\centering
\includegraphics[width=1.0\textwidth]{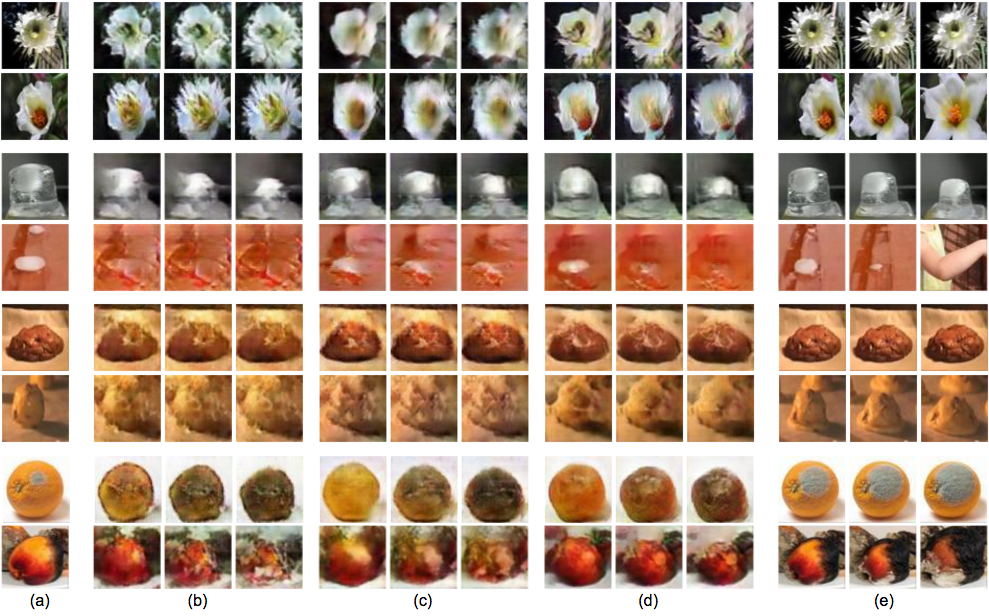}
\caption{Recurrent generation results for: Blooming (rows 1-2), Melting (rows 3-4), Baking (rows 5-6) and Rotting (rows 7-8). Input(a),  p\_mse+adv(b), p\_g\_mse+adv(c), p\_g\_mse+adv+ft(d), Ground truth frames(e).}
%\vspace{-.3cm}
\label{fig:recurrent_gen}
\end{figure}

%\vspace{-.3cm}
\subsection{Recurrent generation}
\label{recurrentexp}
%\vspace{-.1cm}
For our recurrent generation task, we want to train a model to generate
multiple future states of an object given a single input frame. Due to limited
data we recursively generate 4 time steps. During
training, we sample groups of frames from timelapse videos. Each group contains
5 frames, the first being the input, and the rests having 0, 0.1, 0.2, 0.3
degree intervals from the input. As in the previous tasks, the reconstruction
outputs are used for quantitative evaluation. 

We train the models separately for the 4 transformation categories. The models for the {\bf p\_mse+adv} and {\bf p\_g\_mse+adv} methods are trained for 8500 iterations on timelapse data from scratch. For the {\bf p\_g\_mse+adv+ft} method, models are pre-trained on the reconstruction dataset for 5000 iterations then fine-tuned for another 5500 iterations. Fig.~\ref{fig:recurrent_gen} shows prediction results (outputs of $2^{nd}, 3^{rd}$ and $4^{th}$ time steps) of the three methods. Table.~\ref{table:reconstruction} (cols 8-10) shows the reconstruction evaluation.

% table for human eval
\begin{table} [t]
\small{
\begin{center}
\begin{tabular}{|l |c |c| c| c || c| c| c ||c| c| c|}
\cline{2-11}
\multicolumn{1}{c|}{}&\multicolumn{4}{c||}{Pairwise}&\multicolumn{3}{c||}{Two stack}&\multicolumn{3}{c|}{Recurrent}\\
\cline{2-11}
\multicolumn{1}{c|}{}& BL & ADV & Grad & Ft  & ADV & Grad & Ft & ADV & Grad & Ft\\
\hline
Blooming  & 0.1320 & 0.2300& 0.2880 & 0.3500& 0.3080 & 0.3240 & 0.3680 & 0.3320 & 0.2860&0.3820\\
Melting& 0.1680 & 0.2520& 0.2760 & 0.3040 & 0.3400 & 0.3120 & 0.3480 & 0.3180 & 0.3220&0.3600\\
Baking & 0.1620 & 0.2600& 0.2840 & 0.2940 & 0.3120 & 0.3200 & 0.3680 & 0.2780 & 0.3540&0.3680\\
Rotting & 0.1340 & 0.2020 & 0.2580 & 0.4060 & 0.3040 & 0.2640 & 0.4320 & 0.2940 & 0.2900&0.4160\\
Average & 0.1490 & 0.2360 & 0.2765 & {\bf0.3385} & 0.3160 & 0.3050 & {\bf0.3790} & 0.3055 & 0.3130&{\bf0.3815}\\
\hline
\end{tabular}
\end{center}
\caption{Human evaluation results: BL stands for p\_mse method, ADV (p\_mse+adv), Grad(p\_g\_mse+adv) and Ft(p\_g\_mse+adv+ft)}.
%\vspace{-.5cm}
}
\label{table:humanexp}
\end{table}

\begin{figure} [t]
\centering
\includegraphics[width=0.9\textwidth]{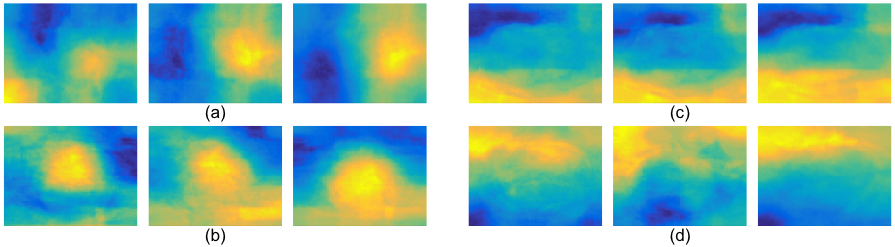}
\caption{Visualization results of learned transformations: x axis flows of blooming (a), y axis flows of melting (b), y axis flows of baking (c) and y axis flows of rotting (d) }
%\vspace{-.3cm}
\label{fig:visual}
\end{figure}

%\vspace{-.3cm}
\section{Additional experiments}
\label{addexp}
%\vspace{-.1cm}

%\vspace{-.1cm}
%\subsection{Human evaluation}
%\label{human}
%\vspace{-.1cm}

%This means that directly comparing future
%predictions to ground truth frames may not be a useful evaluation method. 

\smallskip
\noindent {\bf Human Evaluations:}
As previously described, object future state prediction is sometimes not well
defined. Many different possible futures may be considered
reasonable to a human observer.  Therefore, we design human experiments to
judge the quality of our generated future states.
For each transformation category and generation task, we
randomly pick 500 test cases. Human subjects are shown one
(or two for two stack generation) input image and future images generated by
each method (randomly sorted to avoid biases in human selection). Subjects are
asked to choose the image that most reasonably shows the future object state.

Results are shown in Table 3, where numbers indicate
the fraction of cases where humans selected results from each method as the
best future prediction. For pairwise generation (cols 2-5), the 
{\bf p\_g\_mse+adv+ft} method achieves the most human preferences, while the
baseline performs worst by a large margin. For both two stack generation (cols
6-8) and recurrent generation (cols 9-11), {\bf p\_mse+adv} and 
{\bf p\_g\_mse+adv} are competitive, but again making use of pre-training plus
fine-tuning obtains largest number of human preferences.

%\vspace{-.1cm}
%\subsection{Visualization of the transformation}
%\label{visual}
%\vspace{-.1cm}
\noindent {\bf Image retrieval:}
We also add a simple retrieval experiment on Pairwise generation results using pixelwise similarity. We count retrievals within reasonable distance (20\% of video length) to the ground truth as correct, achieving average accuracies on top-1/5 of {\bf p\_mse+adv}: 0.68/0.94; {\bf p\_g\_mse+adv}: 0.72/0.95; and {\bf p\_g\_mse+adv+ft}: 0.90/0.98.

\noindent {\bf Visualizations:}
Object state transformations often lead to physical changes in the shape of an
object. To further understand what our models have learned, we provide some simple visualizations of 
motion features computed on generated
images. Visualizations are computed on results of the {\bf p\_g\_mse+adv+ft}
recurrent model since we want to show the temporal
trends of the learned transformations. For each testing case, we compute 3
optical flow maps in the x and y directions between the input image and the
second, third, and fourth generated images.  We cluster 
each using kmeans (k=4). Then, for each cluster, we average the optical flow maps
in the x and y directions. 

%We expect that for different state transformations, we will be able to observe
%interpretable and reasonable motion patterns in either the x or y flow maps
%which are consistent with our common knowledge about the transformation.

Fig.~\ref{fig:visual} shows the flow visualization: (a) is the x axis flow
for the blooming transformation. From the visualization we observe the trend of
the object growing spatially. (b) shows the y axis flows for the
melting transformation, showing the object shrinking in the y direction. (c) shows baking, consistent with the object inflating up and down. For
rotting (d), we observe that the upper part of the object inflates with mold or
shrinks due to dehydration.

% The motion of melting transformation usually happen in vertical direction. As the figures show the area of meting are larger and larger in y axis; and up right shows the u axis flows of baking category. They are consistent with the common knowledge that after baking, objects will inflate up and down. And for most testing case the objects are put in the oven and located in the bottom position, thus the movement happen in lower part of the frame; down right is the y axis flows of rotting transformation. Contrary with baking, while rotting, the upper part of the objects start to inflated with mould or shrink because of dehydrating.

%%%%%%%%%%%%%%%%%%%%%%%%%%%%%%%%%%%%%%%%%%%%%%%%%%%%

%\vspace{-.3cm}
\section{Conclusions}
%\vspace{-.1cm}

In this paper, we have collected a new dataset of timelapse videos depicting
temporal object transformations. Using this dataset, we have trained effective
methods to generate one or multiple future object states.  We evaluate each
prediction task under a number of different loss functions and show
improvements using adversarial networks and pre-training. Finally, we provide
human evaluations and visualizations of the learned models. Future work
includes applying our methods to additional single-object transformations and
to more complex transformations involving multiple objects.

\bibliographystyle{splncs}
\bibliography{egbib}
\end{document}